\documentclass[twoside,11pt]{article}

\pdfoutput=1

\usepackage{graphicx,curves}
\usepackage{amsmath}
\usepackage{amssymb}
\usepackage{fullpage}
\usepackage{theapa}
\usepackage{times}
\usepackage{comment}

\newtheorem{df}{Definition}
\newtheorem{ax}{Axiom}

\newtheorem{example}{Example}
\newtheorem{theorem}{Theorem}

\newtheorem{col}{Corollary}

\newcommand{\nin}{\noindent}

\newcommand{\bt}{\begin{theorem}\em}
\newcommand{\et}{\end{theorem}}

\newcommand{\bea}{\begin{eqnarray}}
\newcommand{\eea}{\end{eqnarray}}
\newcommand{\beas}{\begin{eqnarray*}}
\newcommand{\eeas}{\end{eqnarray*}}

\newcommand{\bdf}{\begin{df}\em}
\newcommand{\edf}{\end{df}}

\newcommand{\bax}{\begin{ax}\em}
\newcommand{\eax}{\end{ax}}

\newcommand{\bex}{\begin{example}\em}
\newcommand{\eex}{\end{example}}

\newcommand{\ben}{\begin{enumerate}}
\newcommand{\een}{\end{enumerate}}
\newcommand{\ie}{\item}

\newcommand{\dist}{\operatorname{dist}}

\renewcommand{\min}{\operatornamewithlimits{min}\limits}
\renewcommand{\max}{\operatornamewithlimits{max}\limits}

\newcommand{\gTrap}{$\gamma$-Trap~}
\newcommand{\bgamma}{\bar \gamma}

\newcommand{\Rplusz}{\mathbb R^\oplus}
\newcommand{\Rplus}{\mathbb R^+}
\renewcommand{\top}{\operatorname{top}}
\newcommand{\lb}{\left[}
\newcommand{\rb}{\right]}

\numberwithin{equation}{section}
\numberwithin{theorem}{section}
\numberwithin{lemma}{section}
\numberwithin{col}{section}
\numberwithin{figure}{section}

\begin{document}

\title{On Backtracking in Real-time Heuristic Search}

\author{%
Valeriy K. Bulitko \\
\small Centre for Science, Athabasca University, 1 University Drive \\ 
\small Athabasca, Alberta, T9S 3A3, CANADA \\
\small {\tt valeriyb@athabascau.ca} \and
Vadim Bulitko \\
\small Department of Computing Science, University of Alberta\\
\small Edmonton, Alberta, T6G 2E8, CANADA \\
\small {\tt bulitko@ualberta.ca}}

\maketitle

\begin{abstract}

Real-time heuristic search algorithms are suitable for situated agents that need to make their decisions in constant time. Since the original work by Korf nearly two decades ago, numerous extensions have been suggested. One of the most intriguing extensions is the idea of backtracking wherein the agent decides to return to a previously visited state as opposed to moving forward greedily. This idea has been empirically shown to have a significant impact on various performance measures. The studies have been carried out in particular empirical testbeds with specific real-time search algorithms that use backtracking. Consequently, the extent to which the trends observed are characteristic of backtracking in general is unclear. In this paper, we present the first entirely theoretical study of backtracking in real-time heuristic search. In particular, we present upper bounds on the solution cost exponential and linear in a parameter regulating the amount of backtracking. The results hold for a wide class of real-time heuristic search algorithms that includes many existing algorithms as a small subclass.

\smallskip \nin {\bf Keywords:} real-time heuristic search, agent-centered search.
\end{abstract}

\section{Introduction}\label{sec:introduction}

In this paper we study the problem of {\em agent-centered real-time heuristic search}~\cite{Koenig:00c}. The distinctive property of such search is that an agent must repeatedly plan and execute actions within a constant time interval that is independent of the size of the problem being solved. This restriction severely limits the range of applicable algorithms. For instance, static search algorithms \cite<e.g., A* of>{AStar}, re-planning algorithms \cite<e.g., D* of>{Stenz:95}, anytime algorithms \cite<e.g., ARA* of>{Likhachev:04} and anytime re-planning algorithms \cite<e.g., AD* of>{Likhachev:05} cannot guarantee a constant bound on planning time per action. LRTA* provides such guarantees by planning only a few actions at a time and updating its heuristic function, but the solution quality can be poor~\cite{Korf-AIJ,ishi:92a}.

As a motivating application, consider navigation in gridworld maps in commercial computer games. In such games, an agent can be tasked to go to any location on the map from its current location. The agent must react quickly to the user's command regardless of the map's size and complexity. Consequently, game companies impose a time-per-action limit on their pathfinding algorithms. As an example, Bioware Corp., a major game company, limits planning time to $1$-$3$ ms for all pathfinding units (and there can be many units planning simultaneously). 

The original real-time search algorithms, RTA* and LRTA*, form a local search space (LSS) around the agent's current state. Then they greedily take an action toward the most promising state on the frontier of the LSS. A large number of subsequent real-time heuristic search algorithms have followed this canon~\cite<e.g.,>{RussellWefald-book,FALCONS,ShimboIshida03,Koenig:04a,Hernandez:05b,Hernandez:05a,Koenig:06aamas,Rayner:07-plrta,Bulitko:07-jair,bulitko08-dlrta-jair}. Arguably, the most radical departure was an introduction of the so-called backtracking moves by~\citeA{Shue:93a,Shue:93b,Shue:01}. Their impact on performance of real-time heuristic search and, in particular the cost of the solution the agent finds, has been studied mostly empirically~\cite{Shue:93a,Shue:93b,Shue:01,Bulitko:04a,Bulitko:05d,sverrir:06b}. As a result, it is unclear to what extent the reported findings and trends are specific to the particular algorithms and/or to the testbed environments used. 

The contribution of this paper is an entirely theoretical investigation of effects of backtracking on real-time search performance. We describe a theoretical framework that generalizes a broad class of existing real-time search algorithms. We show that in the worst case, solution cost can be exponential in the parameter controlling the amount of backtracking. We then identify a special case that affords linear solution cost. Because we consider real-time heuristic search on general graphs, the results of our study are domain-independent and, thus, broadly applicable.

The rest of the paper is organized as follows. We first informally review the pioneering LRTA* algorithm and introduce the notion of backtracking in Section~\ref{sec:lrta}. The search problem and performance metrics are formally defined in Section~\ref{sec:notation}. Section~\ref{sec:framework} introduces our framework of real-time search.  We then use the framework to derive properties responsible for exponential (Section~\ref{sec:me}) and linear (Section~\ref{sec:ml}) solution cost. We review existing theoretical work in Section~\ref{sec:related-work-ft-analysis}. The paper is concluded with a discussion of limitations and future work directions. Note that there are no proofs in this version of the paper. We are working on polishing their presentation for a future version.

\section{Backtracking in Real-time Heuristic Search}\label{sec:lrta}

To begin, we present the original real-time search called Learning Real-Time A* or LRTA*~\cite{Korf-AIJ} that constitutes the core of most modern real-time search algorithms. In the current state $s$, LRTA* with a lookahead of one considers the immediate neighbors (lines 4-5 in Figure~\ref{fig:LRTA-simple}). For each neighbor state, two values are computed: the distance of getting there from the current state (henceforth denoted by $g$) and the heuristic estimate $h$ of the distance to the closest goal state from the neighbor state. LRTA* then travels to the state with the lowest $f = g + h$ value (line 7). Then it updates the heuristic value of the current state if the minimum $f$-value is higher (line 6). The process repeats until a goal state is reached.

\begin{figure}[htbp]
\hrule\smallskip
\textbf{LRTA*}

\vspace{1mm}

\begin{tabular} {p{0.2cm}l}
1 & initialize the heuristic: $h \leftarrow h_0$ \\
2 & reset the current state:  $s \leftarrow s_\text{start}$ \\
3 & {\bf while} $s \not \in S_g$ {\bf do} \\
4 & \hspace{2mm} generate successor states of state $s$ \\
5 & \hspace{2mm} among them find the state $s'$ with the lowest $f = g + h$ \\
6 & \hspace{2mm} {\bf if} $f(s') > h(s)$ \\
7 & \hspace{4mm} update $h(s)$ to $f(s')$ \\
8 & \hspace{2mm}  {\bf end if} \\
9 & \hspace{2mm} execute an action to get to $s'$ \\
10 & {\bf end while} \\
\end{tabular}

\smallskip\hrule\vspace{-0.1cm}
\caption{The LRTA* algorithm with a lookahead of one.}\label{fig:LRTA-simple}
\vspace{-0.2cm}
\end{figure}

This paper analyzes the role of backtracking moves which were introduced in an algorithm called Search and Learning A*, or SLA*~\cite{Shue:93a,Shue:93b}. The SLA* is based on the LRTA* algorithm with a lookahead of one we described above with one notable difference. Namely, whenever the heuristic value of the current state is updated (line 7 in Figure~\ref{fig:sla}), the algorithm returns to the previous state (line 8). Otherwise (i.e., when there is no learning), SLA* proceeds to the most promising successor state using the same rule as LRTA* (line 10). Naturally, all backtracking moves incur the travel cost as do regular (forward) moves.

\begin{figure}[t]
\vspace{0.7cm}
\hrule\smallskip
\textbf{SLA*}

\vspace{1mm}

\begin{tabular} {p{0.2cm}l}
1 & initialize the heuristic: $h \leftarrow h_0$ \\
2 & reset the current state:  $s \leftarrow s_\text{start}$ \\
3 & {\bf while} $s \not \in S_g$ {\bf do} \\
4 & \hspace{2mm} generate successor states of state $s$ \\
5 & \hspace{2mm} among them find the state $s'$ with the lowest $f = g + h$ \\
6 & \hspace{2mm} {\bf if} $f(s') > h(s)$ \\
7 & \hspace{4mm} update $h(s)$ to $f(s')$ \\
8 & \hspace{4mm} execute an action to return to the previous state \\
9 & \hspace{2mm} {\bf else} \\
10 & \hspace{4mm} execute an action to get to $s'$ \\
11 & \hspace{2mm} {\bf end if} \\
12 & {\bf end while} \\
\end{tabular}

\smallskip\hrule\vspace{-0.1cm}
\caption{The SLA* algorithm.}\label{fig:sla}
\vspace{-0.2cm}
\end{figure}

The backtracking mechanism provides the agent with two opportunities: (i) to update the heuristic value of the previous state and (ii) possibly select a different action in the previous state. An alternative scheme would be to update heuristic values of previously visited states in memory (i.e., without changing the agent's current state and incurring action cost). This is the approach used by some other real-time search algorithms~\cite{Hernandez:05a,sverrir:06b} and it does not give the agent the opportunity to select a different action.

\section{Definitions and Notation}\label{sec:notation}

In this section we axiomatically introduce the  problem and the formal notation used throughout the rest of the paper.

\bdf A {\em search space} is defined by a connected directed weighted graph $G = (S,E,w)$, and a subset of absorbing (goal) states $S_g \subset G$. Elements of $S$ are states of the search problem and the vertices in the graphs. $E$ is the set of directed edges: $E \subset S \times S$ defining the transitions in the state space. Each edge/transition originating in the state $s$ corresponds to an action the search agent can take in $s$. The function $w : E \to \{ \varepsilon n \mid n \in \Bbb N, n>0\}$ specifies the edge weights (i.e., action costs). Here $\varepsilon \in \Bbb R^+$ is a positive constant. Henceforth, $\Bbb R^+$ is a set of positive real numbers and $\Bbb R^\oplus = \Bbb R^+ \cup \{0\}$. \edf Note that unlike the standard definition \cite<e.g.,>{Bulitko:05d}, in this paper we consider a more general case of a possibly infinite search space and multiple goal states.

\bdf {\em Distance} $\dist(s_1,s_2), s_1, s_2 \in S$ is defined as the minimum cumulative weight of a path originating in $s_1$ and ending in $s_2$. Such a path is called the {\em shortest path}. We generalize this definition for arbitrary sets of states $S', S'' \subset S$ as follows:  $\dist(S',S'') = \min_{s' \in S', s'' \in S''} \dist(s',s'')$. Thus, $\forall s_1, s_2 \in S, \ \dist(s_1,s_2) = \dist(\{s_1\},\{s_2\})$. {\em Distance to goal} of the state $s$ is defined as $h^*(s) = \dist(s,S_g) = \min_{s_g \in S_g} \dist(s,s_g)$. \edf

\bdf {\em Edge-distance} $\|s,s'\|$ is the minimum number of edges among all shortest paths between the states $s$ and $s'$. For any $S', S'' \subset S$, we define $\|S',S''\| = \min_{s' \in S', s'' \in S''} \|s',s''\|$. \edf

\bdf A {\em heuristic search problem} is defined by the search space $G(S,E,w)$, the goal subset $S_g$, an initial heuristic $h_\text{init}$, and an initial state $s_0$. The initial heuristic function $h_\text{init}$ (e.g., Manhattan distance) is a mapping from $S$ to $\{ \varepsilon n \mid n \in \Bbb N, n \ge 0\}$.
The heuristic $h_\text{init}(s)$ is an estimate of $h^*(s)$ and is available to the agent. A heuristic is called {\em $\theta$-admissible} if $\forall s \in S  \lb h(s)\le\theta h^*(s) \rb$; $\theta$ is a positive constant. A heuristic is called {\em consistent} if for any two states $a$ and $b$, $|h(a) - h(b)|$ does not exceed $\dist(a,b)$.\edf

\bdf A heuristic search agent operates as follows. It starts in the initial state $s_0$ and traverses the graph by taking the actions (i.e., directed edges of the graph) until it enters a goal state.  \edf In this paper we do not consider resetting the agent back to its start state upon reaching a goal state and having it solve the problem again. In other words, we are concerned with the first solution only and do not consider the learning process over multiple trials (known as convergence).

\bax\label{ax:sp} The search problem satisfies the following conditions for some $\theta \in \Bbb R, \theta > 0$: \bea
\forall s\in S \ \exists s'\in S_g \lb \dist(s,s')<\infty \rb; \label{eq:G1} \\
\forall s\in S \lb|\{s'|(s,s')\in E\}|<\infty \rb; \label{eq:G2} \\
\forall s\in S \lb h_\text{init}(s)\le\theta \dist(s,S_g) \rb . \label{eq:G4}
\eea
\eax
Condition~\ref{eq:G1} postulates that a goal state is reachable from every state. Next, condition~\ref{eq:G2} states that the number of actions available in any state is finite (i.e., each vertex of the search graph has a finite degree). Finally, condition~\ref{eq:G4} stipulates that the initial heuristic $h_\text{init}$ is $\theta$-admissible. These conditions are needed to ensure completeness (defined below) of the algorithms covered by our framework in the next section. 

\bdf\label{def:completeness} {\bf (Completeness)}. A search algorithm is {\em complete} for a search problem if and only if it necessarily reaches a goal state after a finite number of edge traversals. \edf

\section{A Framework of Learning Real-time Heuristic Search}\label{sec:framework}

This section presents a framework of real-time heuristic search which we subsequently use in our analysis. We introduce the framework axiomatically and then illustrate it with examples.

\subsection{Search Framework}

\bdf We call $S' \subset S$ a {\em separating set}  for the state $s$ if every shortest path from $s$ to every goal state $s_g \in S_g$, if it exists, passes
through $S'$. \edf From Condition~\ref{eq:G1} of Axiom~\ref{ax:sp}, it follows that any non-goal state has a non-empty separating set (we can define $S'$ as a set of states from all shortest paths between $s$ and all goal states reachable from $s$).

\bdf Suppose $S' \subset S$ and $s \in S$, then $\mathfrak D(s,S')$ denotes the set of all subsets of $S'$ which happen to be separating sets for $s$. \edf Clearly, for any non-goal state $s$, $\mathfrak D(s,S)$ contains at least one element.

\begin{figure}[htbp]
\begin{center}
\includegraphics[width=6cm]{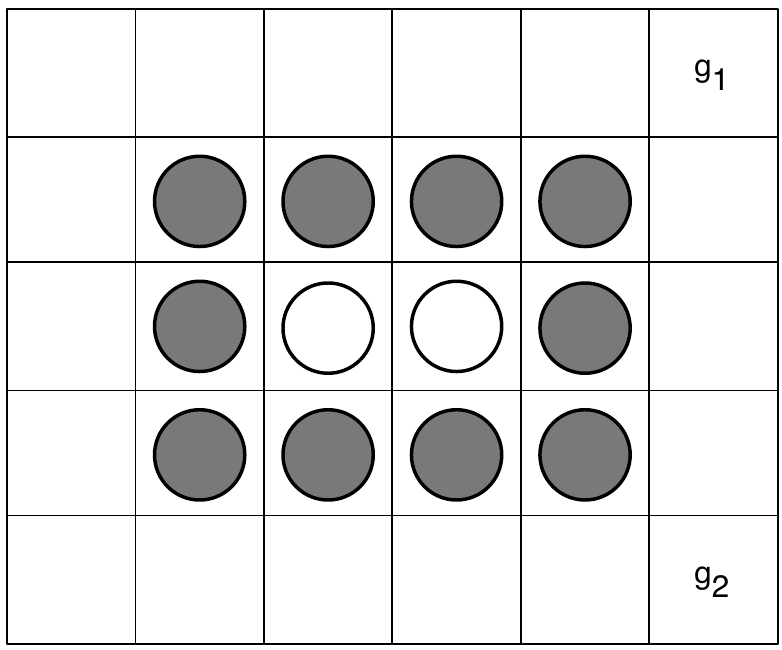}
\caption{A gridworld illustration of Definition~\ref{def:border}. The states in $S$ are $30$ grid cells in the $6 \times 5$ grid world. The two goal states are marked as $g_1$ and $g_2$. States in $\Gamma$ are marked with circles. Light circles represent the inner states of $\Gamma$. Shaded circles form the border $\partial \Gamma$.}\label{fig:border}
\end{center}\vspace{-0.7cm}
\end{figure}

\bdf\label{def:border} We define the {\em border} of any set  $\Gamma \subset S$ as $\partial \Gamma = \{ s \in \Gamma \mid \exists s' \in S \setminus \Gamma$ such that the edge $(s,s') \in E \}$. States in $\Gamma \setminus \partial \Gamma$ are called {\em inner states} of $\Gamma$. \edf Clearly, the border of any set $\Gamma$ that does not contain inner goal states, is a separating set for an inner state $s$: $S_g \cap \Gamma \subset \partial \Gamma \implies \forall s \left[ s \in \Gamma \setminus \partial \Gamma \implies \partial \Gamma \in \frak D(s,\Gamma) \right]$. An example is found in Figure~\ref{fig:border}.

\begin{figure}[t]
\hrule\smallskip
\textbf{Search algorithm $\pi(\theta,T)$}

\vspace{1mm}

\begin{tabular} {p{0.2cm}l}
1 & reset the cycle counter: $t \leftarrow 0$ \\
2 & initialize the heuristic: $h_0 \leftarrow
h_\text{init}$ \\
3 & reset the stack:  $\sigma_0 \leftarrow [s_0]$ \\
4 & reset the learning amount: $u_0 \leftarrow 0$ \\
5 & {\bf while} $\top(\sigma_t) \not \in S_g$ {\bf do} \\
6 & \hspace{4mm} generate a local search space $\Gamma(\sigma_t)$ \\
7 & \hspace{4mm} compute the heuristic
weight $\gamma(\sigma_t)$ and the new heuristic
$h_{t+1}$ \\
8 & \hspace{4mm} update the learning amount $u_{t+1}$ \\
9 & \hspace{4mm} take the action and compute the new
stack $\sigma_{t+1}$ \\
10 & \hspace{4mm} advance the cycle counter:
$t \leftarrow t + 1$ \\
11 & {\bf end while} \\
\end{tabular}

\smallskip\hrule
\caption{The real-time search framework analyzed in this paper.}\label{fig:framework}\vspace{-0.3cm}
\end{figure}

We will now define a family of algorithms covered by our analysis. 

\bdf\label{def:framework} The search framework in Figure~\ref{fig:framework} implicitly defines a class of search algorithms. Any member of this class will be referred to as search algorithm $\pi(\theta,T)$ and is invoked by a search agent at discrete time steps. \edf 

The fundamental part of the family of algorithms is the concept of stack. The stack is used to represent the path from the start state to the agent's current state. By analyzing stack's evolution, we will be able to formulate bounds on algorithm's performance. We define the stack notation below and then walk through the framework line by line.

\bdf\label{def:stack} The stack is a first-in-last-out data structure that maintains the path found by the agent from its start state to its current state. The notation  $\sigma_t = [s_1 \dots s_n]$ means that the stack $\sigma$ contains states $s_1, \dots, s_n$ at time $t$; state $s_1$ is the start state, state $s_n$ is the current state. The top of the stack $\sigma = [s_1 \dots s_n]$ is $s_n$ and is denoted by $\top(\sigma)$. Furthermore, if $\sigma$ and $\sigma'$ are stacks then their concatenation $\sigma\sigma'$ is a stack. For brevity, we will use this notation for a concatenation of a stack with a single state (e.g., $\sigma s$). If $\sigma = s_1 \dots s_n$ then $\sigma |^k = s_1 \dots s_k$ and $\sigma |_k = s_{n-k+1} \dots s_n$ (i.e., the first and the last $k$ elements of an $n$-element stack respectively). Notation $\sigma|^b_a, a<b$ represents stack elements $\sigma_a, \dots, \sigma_b$. Finally, $|\sigma|$ stands for the number of elements in the stack $\sigma$.\edf

During the initialization (lines 1 - 4 in Figure~\ref{fig:framework}), the cycle counter is reset to $0$ and the start state $s_0$ is  pushed onto the empty stack $\sigma_0$. The heuristic function at time $0$ (denoted by $h_0$) is set to the initial heuristic $h_\text{init}$. Finally, a real-valued counter, called {\em learning amount} $u_0$, is cleared. On each cycle (lines 6 -- 10), the algorithm goes through planning, learning, and execution phases. In the planning phase (line 6), the algorithm examines the states around the current state. The set $\Gamma(\sigma_t)$ of the neighboring states examined is henceforth called {\em local search space}. $\Gamma(\sigma_t)$ can be defined in many ways. The results in this paper apply to any definition of $\Gamma(\sigma_t)$ that satisfies the forthcoming Axiom~\ref{ax:as}.

Learning happens in lines 7 and 8 where the algorithm computes the heuristic weight $\gamma(\sigma_t)$, the new heuristic function $h_{t+1}$ and the new learning amount $u_{t+1}$. Again, in the interest of covering as many heuristic update rules as possible, we do not specify a particular rule. Any learning rule  satisfying Axiom~\ref{ax:as} is covered by our analysis.

In line 9, the agent executes its move by updating its stack from $\sigma_t$ to $\sigma_{t+1}$. We say that the agent takes a {\em forward move} if $|\sigma_{t+1}| = |\sigma_t| + 1$ (i.e., the stack grows) and a {\em backward move} if $|\sigma_{t+1}| = |\sigma_t| - 1$ (i.e., the stack shrinks). 

\begin{figure}[htbp]
\begin{center}
\includegraphics[width=9cm]{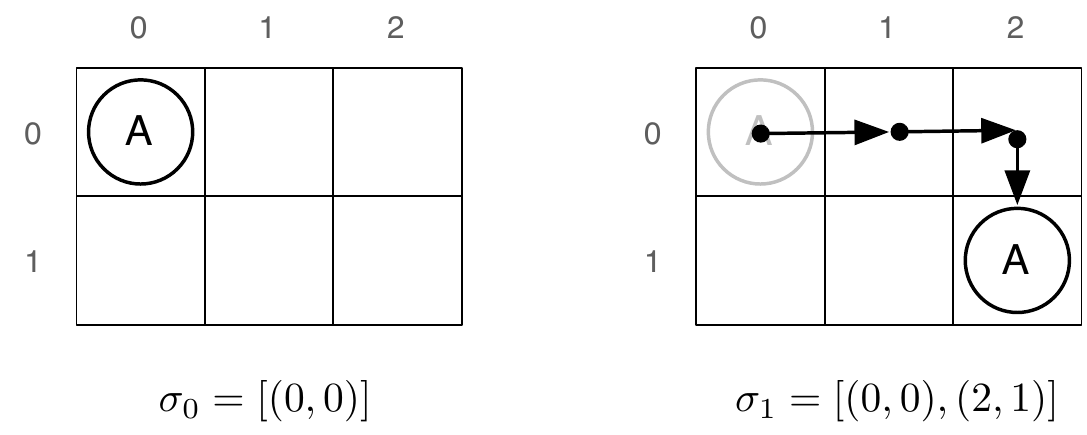}
\caption{{\bf Left:} the initial state and the corresponding stack $\sigma_0$. {\bf Right:} the next state after three moves are taken and the resulting stack $\sigma_1$.}\label{fig:macroMoves}
\end{center}\vspace{-0.7cm}
\end{figure}

Note that real-time search algorithms can execute several actions per a single planning cycle. In our framework, the stack $\sigma_t$ stores only the states in which planning was carried out. As an illustration, consider the gridworld example in Figure~\ref{fig:macroMoves}. Suppose a path-finding agent starts out in a state with the coordinates $(0,0)$ and plans three moves -- two right moves followed by a down move. This means that by the time it plans again, it will have visited the states $(1,0), (2,0), (2,1)$. In this paper, we discard the intermediate states $(1,0)$ and $(2,0)$ and update our stack from $\sigma_0 = [(0,0)]$ to $\sigma_1 = [(0,0), (2,1)]$. This simplification is strictly for notational convenience and the discarded states are indeed visited by the agent. 

We increment the cycle counter in line 10.  The loop is terminated as soon as a goal state appears on top of the stack (line 5).

\subsection{Conditions}

The framework in Figure~\ref{fig:framework} defines a broad class of search algorithms. In order to prove completeness and upper bounds on solution, we introduce the following restrictions.

\bax\label{ax:as} Suppose the {\em learning quota}  $T
\in \Rplusz \cup \{ \infty \}$, and the {\em
admissibility weight} $\theta \in \Rplus$ are control
parameters. Then the following conditions are imposed
on the search framework in Figure~\ref{fig:framework}
for any step $t$: \bea
\top(\sigma_t) \not \in S_g \implies \frak D(\top(\sigma_t),
 \Gamma(\sigma_t)) \neq \emptyset  \label{eq:ce1} \\
  0 < \gamma(\sigma_t) \le \bgamma \label{eq:c1a}\\
 \ |\sigma_{t+1}| > |\sigma_t| \implies \exists s \in
 \Gamma(\sigma_t) \ \lb \sigma_{t+1} = \sigma_t  s
  \And \nonumber \right. \\   \left.
  h_{t+1}(\top(\sigma_t)) \ge \gamma(\sigma_t)
  \dist(\gamma(\sigma_t),s) + h_{t+1}(s) \rb \label{eq:c2}
   \\
 \ |\sigma_{t+1}| < |\sigma_t| \implies
 h_{t+1}(\top(\sigma_t)) > h_t(\top(\sigma_t))
 \label{eq:c3} \\
\ \forall s \in S \ \lb h_t(s) \le \theta h^*(s)
\rb \label{eq:c4} \\
\ \forall s \in \Gamma(\sigma_t) \ \lb h_{t+1}(s)
\ge h_t(s) \rb \And \forall s \not \in \Gamma(\sigma_t)
\ \lb h_{t+1}(s) = h_t(s) \rb \label{eq:c5} \\
\text{let   }\  \nabla_t = \begin{cases}
\left\{ s_0 \right\}, & |\sigma_{t+1}| > |\sigma_t|, \\
\left\{ s_0, \top(\sigma_t) \right\}, & |\sigma_{t+1}|
< |\sigma_t|,
\end{cases} \nonumber \\ \text{ then   }\
T \ge u_{t+1} \text{ and } u_{t+1} = u_t + \underset{\substack{s \in
\sigma_{t+1}, \\ s \not\in \nabla_{t+1}}} {\sum}
\lb h_{t+1}(s) - h_t(s) \rb
\label{eq:c6} \eea \nin The meaning of the conditions is as follows: \begin{description}
\item[Condition~\ref{eq:ce1}] requires that the lookahead search space $\Gamma(\sigma_t)$ contain a separating set for the current non-goal state $\top(\sigma_t)$.
\item[Condition~\ref{eq:c1a}] places an upper bound on the dynamically selected heuristic weight $\gamma(\sigma_t)$.
\item[Condition~\ref{eq:c2}] mandates that whenever the stack grows, the new stack is produced by pushing a single state $s$ from the local search space $\Gamma(\sigma_t)$ on the previous stack. Additionally, the updated heuristic value of current state $\top(\sigma_t)$ must be lower-bounded by the distance from the current state $\top(\sigma_t)$ to the new state $s$ weighted by $\gamma(\sigma_t)$ plus the distance from $s$ to goal, as estimated by $h_{t+1}(s)$.
\item[Condition~\ref{eq:c3}] requires the heuristic value of the current state to strictly increase whenever backtracking (i.e., popping the stack) occurs.
\item[Condition~\ref{eq:c4}] requires the heuristic to be $\theta$-admissible at all times.
\item[Condition~\ref{eq:c5}] postulates that the heuristic function cannot change outside of the lookahead search space  $\Gamma(\sigma_t)$ and it can only increase inside the space.
\item[Condition~\ref{eq:c6}] states that  the sum of all increments in the heuristic function over all states except the exclusion set $\nabla_t$ does not exceed the learning quota $T$. It also requires that the learning amount increases from $u_t$ to $u_{t+1}$ by the cumulative increment in the heuristic function from time $t$ to time $t+1$. The exclusion set $\nabla_t$ always contains the start state $s_0$. If the move executed at time $t$ is a backtracking move, it also contains the current state $\top(\sigma_t)$. 
\end{description}
\eax

Conditions~\ref{eq:c2} and~\ref{eq:c6} imply that the agent is allowed to do no more than $T$ learning (i.e., increases to its heuristic) while moving forward. Once the quota is exhausted, it will have to backtrack every time it learns and is allowed to move forward only when its heuristic is locally consistent (i.e., no learning is required). This is in line with existing algorithms SLA*T~\cite{Shue:01} and LRTS~\cite{Bulitko:05d}.

\bdf\label{def:ftsc} We define the {\em solution cost} as the cumulative distances between the consecutive states on the agent's stack upon reaching a goal state. Specifically, suppose that upon reaching a goal state the stack $\sigma_n = [s_0, \dots, s_m]$. Then, the solution cost is $\sum_{j=1}^m \dist(s_{j-1},s_j)$.\edf

Note that solution cost is different from the number of moves on the first trial~\cite{koen:92b,Koenig:93,Koenig:96} or the execution cost~\cite{Bulitko:05d}. Specifically, there are two differences: we do not count take the cost of backtracking moves into account and we use shortest-path costs between states on the stack. As noted above (Figure~\ref{fig:macroMoves}), only states in which the agent conducted planning are put on the stack. Our definition assumes that the path segment computed within a single planning session is optimal (i.e., the actual edge costs add to $\dist(s_{j-1},s_j)$). This is a realistic assumption for many algorithms because either they take a single move per planning session or their moves are guaranteed to be optimal within the local search space expanded at each planning session.

We chose to use solution cost  so defined as the performance measure because it provides an insight into the role of solution stack and backtracking in real-time heuristic search. As a consequence, we are able to derive  upper bounds linear in the control parameter ($T$) as opposed to the tight bounds quadratic in the state space size \cite{koen:92b,Koenig:93,Koenig:96}.

\bdf\label{def:admissibleSearch} An instance $\pi$ of the search framework in  Figure~\ref{fig:framework} is said to be {\em $\theta$-admissible search} if it  satisfies Axiom~\ref{ax:as} for a given value of $\theta$. \edf

Note that Axiom~\ref{ax:as} places no restrictions on the size of local search space $\Gamma(\sigma_t)$. In other words, non-real-time search algorithms that do not guarantee a move in a constant amount of time are covered by our analysis as well.

\subsection{Examples}

We will now illustrate the new notation and definitions with several examples.

\begin{figure}[ht]\vspace{0.3cm}
\begin{center}
\includegraphics[width=8cm]{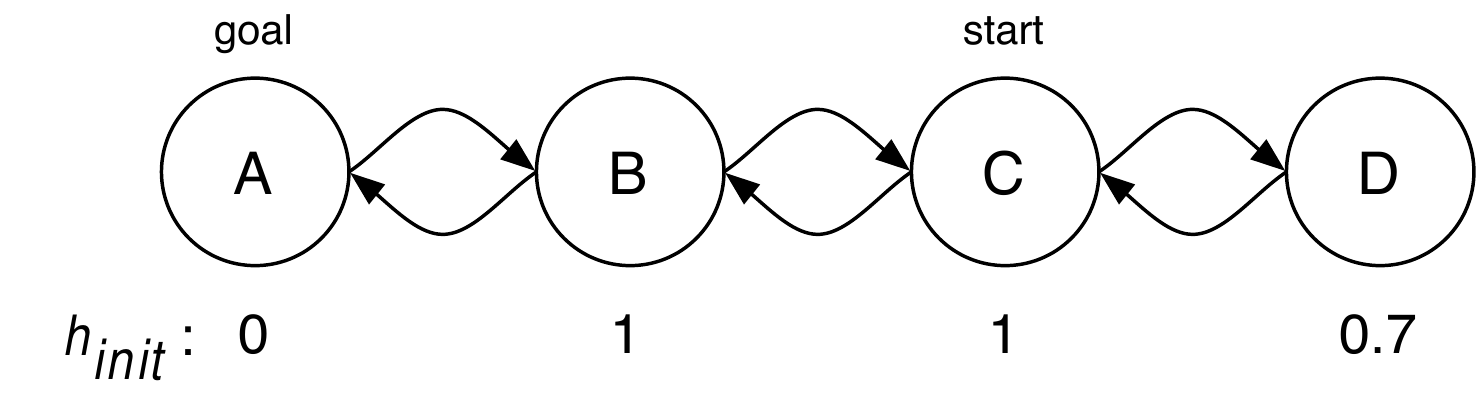}
\caption{A four-state example.}\label{fig:ex1}
\end{center}\vspace{-0.7cm}
\end{figure}

\bex\label{ex:LRTA} {\bf (Korf's LRTA*)}.  Consider a four-state space in Figure~\ref{fig:ex1}. In our notation, the problem is specified as $G = (\{A,B,C,D\}, \{(A,B), (B,C), (C,D), (D,C), (C,B), (B,A)\}, w)$ where the weight function $w$ is $1$ for all six edges. The goal set $S_g = \{A\}$ and the start state $s_0 = C$. The admissibility weight $\theta = 1$ and the learning quota $T= \infty$. The initial heuristic function is defined as follows: $h_\text{init}(A) = 0, h_\text{init}(B) = 1, h_\text{init}(C) = 1, h_\text{init}(D) = 0.7$. Korf's LRTA* with a lookahead of one (Figure~\ref{fig:LRTA-simple}) constitutes a $1$-admissible search and can be represented in our framework as follows:
\ben

\ie the neighborhood $\Gamma(\sigma_t)$  is defined as the immediate neighbors of a current state. Note that there is no left neighbor for state $A$ and no right neighbor for state $D$;

\ie the heuristic weight $\gamma(\sigma_t)$  is set to $1$ for all $\sigma_t$ (thus, $\bgamma = 1$);

\ie the stack always grows by one state --- either the left or the right neighbor of the current state: $|\sigma_{t+1}| = |\sigma_t| +1$ for all $t$. Namely, the agent goes to the immediate neighbor $s$ with the lowest $f$-value: $f(s) = \gamma(\sigma_t) \dist(\top(\sigma_t),s) + h_t(s)$. The ties can be broken in any of the standard ways~\cite<e.g., randomly, systematically, or in a fixed order as done by>{FALCONS};

\ie the heuristic is updated only in the  current state as follows: \bea h_{t+1}(\top(\sigma_t)) = \max \left\{ h_t(\top(\sigma_t)) , \min_{s \in
\Gamma(\top(\sigma_t))} \left( \gamma(\sigma_t) \dist(\top(\sigma_t),s) + h_t(s) \right) \right\}; \eea

\ie the learning amount is computed as  follows: \bea u_0 = 0, \nonumber \\ \forall t \ge 0 \left[ u_{t+1} = u_t + h_{t+1}(\top(\sigma_t)) - h_t(\top(\sigma_t)) \right] . \eea

\een

\nin The following table shows the  heuristic values (with the value of the current state in bold), the contents of the stack $\sigma_t$, the local search space $\Gamma(\sigma_t)$, and the learning amount $u_t$:

\begin{center}
\begin{tabular}{c|cccc|c|c|c}
\hline
$t$ & $h_t(A)$ & $h_t(B)$ & $h_t(C)$ & $h_t(D)$ & $\sigma_t$ & $\Gamma(\sigma_t)$ & $u_t$ \\
\hline
0 & 0 & 1 & \bf 1 & 0.7 & [C] & \{B,D\} & 0 \\
1 & 0 & 1 & 1.7 & \bf 0.7 & [C,D] & \{C\} & 0.7 \\
2 & 0 & 1 & \bf 1.7 & 2.7 & [C,D,C] & \{B,D\} & 2.7 \\
3 & 0 & \bf 1 & 2.0 & 2.7 & [C,D,C,B] & \{A,C\} & 3 \\
4 & \bf 0 & 1 & 2.0 & 2.7 & [C,D,C,B,A] & \{B\} & 3 \\
\hline
\end{tabular}
\end{center}

\nin The  solution cost is $4$ as it is the cost of the path on the final stack: $C \to D \to C \to B \to A$.

\eex

It is straightforward to show that any parameterization of the LRTS algorithm~\cite{Bulitko:05d} constitutes a $\theta$-admissible search in our framework with $\theta$ set to the value of $\gamma$ in LRTS. As a corollary, LRTA*~\cite{Korf-AIJ}, weighted LRTA*~\cite{ShimboIshida03}, SLA*~\cite{Shue:93a,Shue:93b}, and \gTrap~\cite{Bulitko:04a} are covered by our framework as well.
The inclusion is proper. The following example shows that there are $\theta$-admissible policies that are not instances of LRTS.

\bex\label{ex:beyondLRTS} {\bf (Beyond LRTS)}. Consider the 4-state search space and the initial heuristic from Example~\ref{ex:LRTA}. The start state $C$ is called $1$-trap~\cite{Bulitko:04a} with respect to the immediate neighbors since the $f$-value of the left neighbor is $f(B) = \dist(C,B) + h_\text{init}(B) = 2$, the $f$-value of the right neighbor is $f(D) = \dist(C,D) + h_\text{init}(D) = 1.7$ and they both exceed the heuristic value of the start state: $h_\text{init}(C) = 1$. In such cases, the agent may want to expand the search space until the current state is no longer a trap or an upper-bound $d_\text{max}$ is reached. This can be implemented in the framework: \bea
\Gamma(\sigma_t) &=& S(s_c,d) \\
s_c &=& \top(\sigma_t) \\
S(s_c,k) &=& \{ s \in S \mid \|s_c,s\| = k\} \\
d &=& \min \left\{ d_\text{max}, \|s_c, S_g \|, \min \left\{ k \left| \ h_t(s_c) \ge
 \min_{s \in S(s_c,k)} f(s) \right. \right\} \right\} \label{line:dynamicGrowth}\\
f(s) &=& \gamma(\sigma_t) \dist(s_c, s) +
 h_t(s)\label{eq:f}.
\eea \nin Intuitively, $S(s_c,d)$ defines a depth $d$ full-width neighborhood which is guaranteed to be a separating set unless $s_c \in
S_g$. Additionally, we update the heuristic over the entire lookahead space:
\bea\label{eq:maxmins} \forall s \in \Gamma(\sigma_t)  \left[ h_{t+1}(s) = \max\left\{ h_t(s) , \max_{S' \in \frak D(s, \Gamma(\sigma_t))} \ \min_{s' \in S'} f(s') \right\} \right], \eea
\nin where $f(\cdot)$ is defined in Equation~\ref{eq:f}. Note that all updates from $h_t$ to $h_{t+1}$ are done in parallel \cite<called synchronous backups by>{barto95}. The fact that the new value of each $s \in \Gamma(\sigma_t)$ is a minimum $f$-value over a separating set $S' \in \frak D(s, \cdot)$ maintains $\gamma$-admissibility of the heuristic. The distance weight $\gamma(\sigma_t)$ is set to $1$ for all $t$. We finalize the algorithm by setting $T$ to $\infty$ thereby disabling backtracking. The resulting algorithm is $1$-admissible search in the sense of Axiom~\ref{ax:as}. On the other hand, the algorithm goes beyond the LRTS framework due to dynamic search space growth (line~\ref{line:dynamicGrowth}) and multiple heuristic updates (line~\ref{eq:maxmins}).
\eex

In Example~\ref{ex:beyondLRTS} above,  we used the \gTrap's and LRTS' ``max of mins" update rule (Equation~\ref{eq:maxmins}) which maintains $\gamma$-admissibility of the heuristic. The intuition lies with the fact that the minima are sought over separating sets. This means that each minimum is computed over a set of states that an optimal path to every goal state through. Clearly, it is safe to increase the heuristic value of the current node to the $f$-value of any state on an optimal path as it will not violate (weighted) admissibility. Thus, it is safe to increase $h$ of the current state to the minimum $f$ value of {\em any} separating set. As the initial heuristic is $\theta$-admissible, it makes sense to increase it as aggressively as possible (hence the $\max$ in the update rule). Note that the classic LRTA* seeks the minimum of $f$-values over depth $d$ frontier which is a separating set. LRTS and \gTrap are more aggressive and look at a series of frontiers for the depth values of $1, \dots, d$. They then select the highest minimum as the new value of the heuristic value of the current state.

Interestingly, the ``max of mins" update rule is a sufficient but not necessary to preserve $\theta$-admissibility of heuristic function. In Appendix~\ref{sec:max-min-criterion} we show that even more aggressive $\theta$-admissible rules are possible. More importantly, we derive a criterion for $\theta$-admissibility and construct an non-trivial upper-bound on the magnitude of the updates for any $\theta$-admissible learning rule.

\section{Theoretical Analysis}

We start by proving that any $\theta$-admissible search (Definition~\ref{def:admissibleSearch}) is complete.

\bt\label{th:completeness} {\bf (Completeness).} When $T < \infty$ or $|S| < \infty$, any $\theta$-admissible search $\pi$, $\bgamma\le\theta$ starting with a $\theta$-admissible heuristic is complete.

 \et

\begin{col}\em\label{cor_compl} When $T < \infty$ or $|S| < \infty$, any $\bgamma$-admissible search $\pi$ starting with a $\bgamma$-admissible heuristic is complete.
\end{col}

\nin Consequently, in the rest of the paper, we will implicitly  assume that $\theta \ge \bgamma$ so that the search $\pi$ is $\theta$-admissible. This assumption can always be satisfied by increasing $\theta$ to $\bgamma$ if necessary.

\subsection{Exponential Solution Cost}\label{sec:me}

The amount of backtracking an agent performs is determined by inaccuracies of the initial heuristic as well as the learning quota control parameter ($T$). Is there a relation between $T$ and solution cost produced by an algorithm? \citeA{Bulitko:05d} showed that LRTS that prunes out duplicate states on its solution stack has a solution cost linear in $T$ (cf. Section~\ref{sec:related-work-ft-analysis} for more discussion). Until now, it was not known whether this upper bound is specific to LRTS. In this section we show that any $\theta$-admissible search has an upper bound on the  solution cost exponential in $T$. We then show that the bound is tight by demonstrating an example in which the the bound is achieved.

\bt {\bf (Exponential upper-bound)}. \label{th:expUB} There exists a positive integer $\Delta$ such that for any $\theta$-admissible search $\pi$ and any start state $s_0$ the  solution cost is upper-bounded by:
$$
\left(2\theta
\dist(s_0, S_g)+\frac{\Delta \varepsilon}{\ln
2}\right)2^{\frac{T}{\Delta \varepsilon}}.
$$
\et

\bt {\bf (Example of exponential  cost)}. There exists a search problem $G$ and a $1$-admissible search $\pi, T=12m, \bgamma=1$ where $m$ is a positive integer such that the  solution cost is $9 \cdot 2^{m+1}-5$ (i.e., exponential in $T$).
\label{th:expSolCost}\et

\subsection{Linear Solution Cost}\label{sec:ml}

\begin{figure}[ht]
\begin{center}
\includegraphics[width=6.5cm]{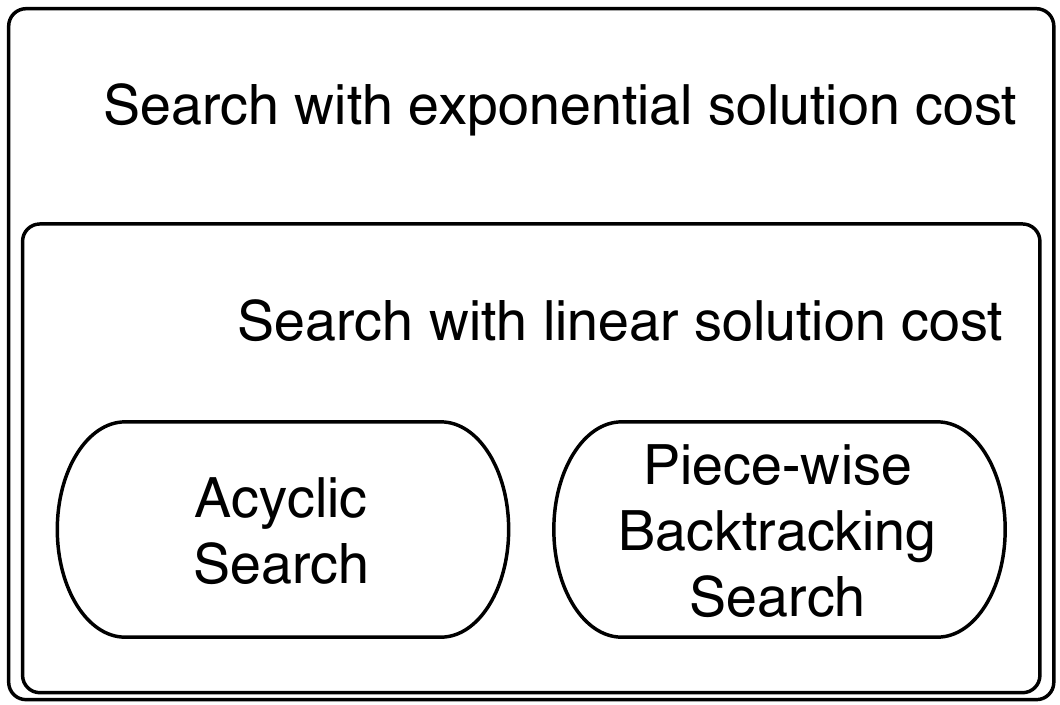}
\vspace{-0.2cm} \caption{Structure of the set of $\theta$-admissible search algorithms.}\label{fig:classes}
\end{center}\vspace{-0.4cm}
\end{figure}

In this section we define a subclass of  the exponential class whose members have  solution cost linear in $T$. We then demonstrate that this class is non-empty by constructively defining two families of policies that belong to the linear subclass (Figure~\ref{fig:classes}).

\bdf {\bf (Linear subclass).}  A search $\pi$ belongs to the linear class if its  solution cost has an upper bound linear in $T$.  Specifically, a $\theta$-admissible search $\pi$ belongs to the linear subclass if there exist non-negative numbers $a,b,c$ (that possibly depend on $\theta,\bgamma$) such that the  solution cost is upper-bounded by $a\dist(s_0,S_g) + bT + c$. \edf

While traversing the state space,  a $\theta$-admissible search can visit a single state more than once. Such re-visits happen either through backtracking moves (the stack shrinks) or forward moves (the stack grows). In the latter case, several copies of the revisited state will be present on the search's path stack at once thereby forming a cycle in the solution.

\bdf\label{def:acyclic} A $\theta$-admissible search  is called {\em acyclic} when its stack never contains multiple occurrences of a state. \edf

The LRTS algorithm of \citeauthor{Bulitko:05d} \citeyear{Bulitko:05d} is an example of acyclic search because it explicitly removes any cycles in its solution.

\bt\label{th:acyclic} Any acyclic $\theta$-admissible search belongs to the linear class. \et

Another example of search with linear solution bound is achieved through limited backtracking as follows.

\bdf\label{def:pwbs} A $\theta$-admissible search $\pi$ is called {\em piecewise backtracking} if for a positive integer $k$ the following conditions hold: \ben

\ie the path stack is divided into a finite number of segments. Each segment except possibly the last (i.e., top-most) segment are exactly $k$-states each. We will denote stack segment $i$ by $\sigma|_{b_i}^{e_i}$ where $s_{b_i}$ is the first state and $s_{e_i}$ is the last state of the stack segment. For all, except possibly the last segment, $e_i - b_i = k-1$.  The first segment begins with the start state: $b_1 = 0$;

\ie within each segment, every increase  in the heuristic function (i.e., when $h_t(s) < h_{t+1}(s)$) results in a backtracking move as long as it does not bring the agent  into the previous segment. In other words, if the current state $s = \top(\sigma_t)$ is the first state of a segment than no move is taken at all when $h_t(s)$ is increased (i.e., the agent stays put);

\ie every time the current stack segment (i.e., the segment containing the top of the stack) grows beyond $k$ states, a new segment is started. Suppose its first state is $\top(\sigma) = s_{b_M}$. Then the following quantity is computed:
\bea\label{eq:tdiscr}
\sum_{i=1}^{M-1}
h(s_{b_{i+1}}) - h(s_{e_{i}}) + \dist(s_{e_{i}},s_{b_{i+1}}).
\eea
Once this quantity exceeds $T$, the segment that was just started is declared final. This means that it can grow beyond $k$ states. As with other segments, each increase in a heuristic value forces the agent to take backtracking move {\em within} the segment. Backtracking is not allowed into the previous segments (i.e., past $s_{b_M}$).

\een \edf

\begin{figure}[ht]
\begin{center}
\includegraphics[width=14cm]{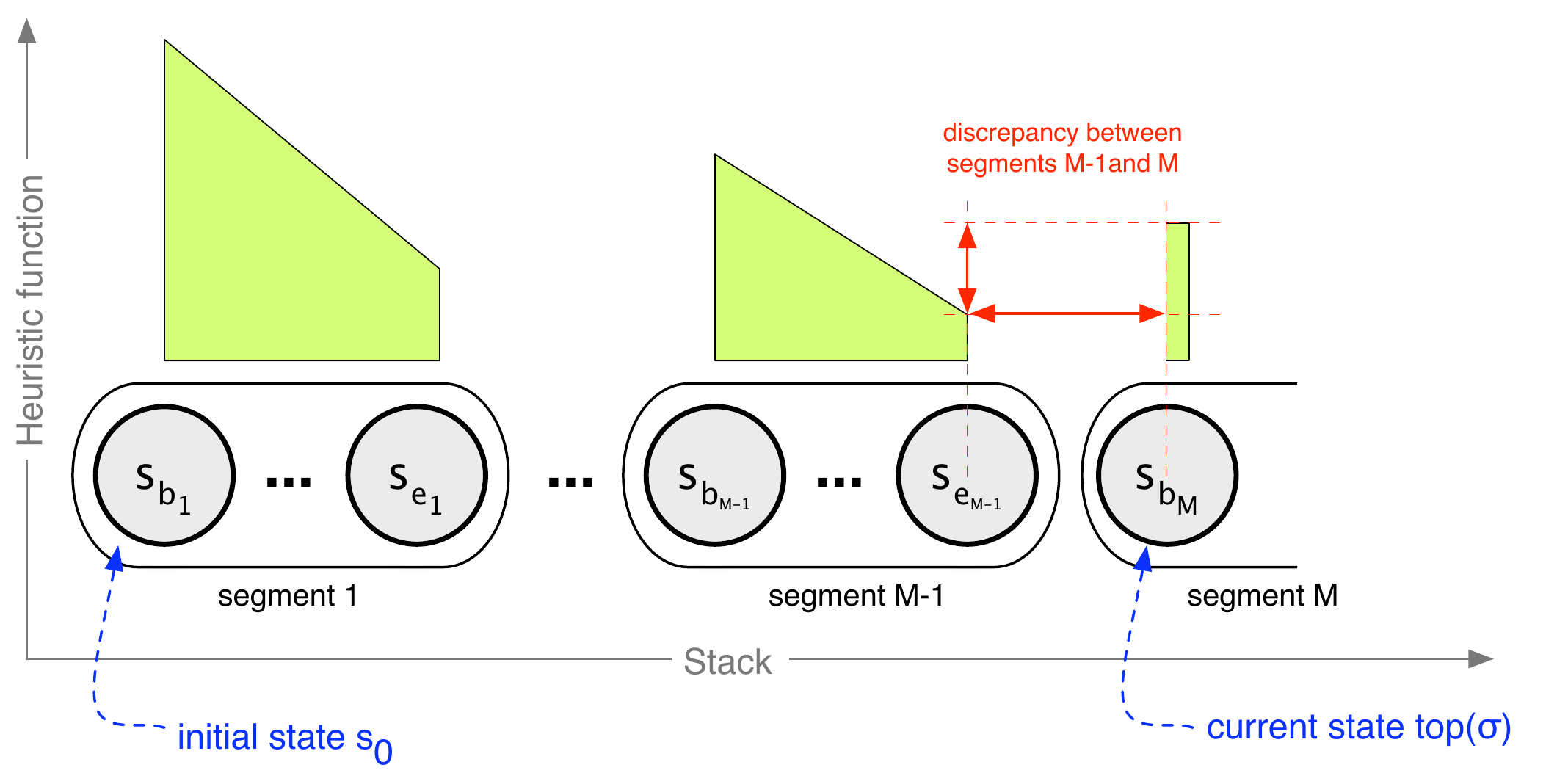}
\vspace{-0.3cm}
\caption{Piecewise backtracking search upon entering its segment $M$.}\label{fig:piecewise}
\end{center}\vspace{-0.3cm}
\end{figure}

Figure~\ref{fig:piecewise} illustrates the definition. It shows the situation precisely upon starting a new segment $M$. The stack runs horizontally, left to right. Heuristic values are shown vertically. The trapezoids have smooth upper edge indicating consistency of the heuristic function on each segment. The
sharp discontinuities between heights of the neighboring trapezoids represent the discrepancies in the heuristic function of the neighboring segments. Together with the distances between the states they form the sum in Equation~\ref{eq:tdiscr}. We schematically show the last of the sum's terms (i.e., $h(s_{b_M}) - h(s_{e_{M-1}}) + \dist(s_{e_{M-1}},s_{b_M})$) with bidirectional arrows.

\bt\label{th:pwbs} A piecewise backtracking search $\pi$ has solution cost linearly upper-bounded by: $$3\theta \dist(s_0,S_g)+2T$$ for any positive integer $k$.
\et

\section{Related Work on Solution Cost Analysis}\label{sec:related-work-ft-analysis}

\citeA{MTS} showed that LRTA* is guaranteed to reach the goal in $O(n^2)$ steps if the state space of size $n$ has no identity actions. As~\citeA{koen:92b,Koenig:93,Koenig:96} point out, this follows from the analysis of MTS if the target's position is fixed~\cite{koen:92a}.

\citeA{koen:92b,Koenig:93,Koenig:96} extended the analysis onto a sub-class of reinforcement learning algorithms of which LRTA* is a special case. They considered state- as well state-action value functions and two schemes of reinforcement: action penalty and goal rewards. For the action-penalty approach, they proved that the time-complexity of the first trial is upper bounded by $O(n^3)$ for Q-learning algorithms and by $O(n^2)$ for value-iteration algorithms. Both upper bounds are tight for zero-initialized heuristics.

It can be shown that SLA* discussed in the previous section finds an optimal path by the time it arrives at the goal state. Furthermore, SLA* will have learnt perfect heuristic values for all states on such a path~\cite{Shue:93a,Shue:93b}. A subsequent algorithm, named SLA*T~\cite{Shue:01,Zamani:01}, introduced a learning quota parameter $T$ which makes SLA*T behave identically to LRTA* (i.e., no backtracking) as long as the overall amount of heuristic updates is under $T$. As soon as this threshold is exceeded, SLA*T starts behaving identically to SLA* (i.e., backtracks on every heuristic update). \citeA{Bulitko:05d} showed that the length of solution found by SLA*T is upper-bounded by $\dist(s_0,S_g) + T$ where $T$ is the learning quota/learning threshold parameter.
This result holds {\em only} when the path being built by SLA*T is processed after every move and all state revisits are removed. The downside of this
requirement is that the pruning operation substantially increases the running time of the algorithm and can require a non-constant amount of time per move.

\section{Future Work}

The analysis in this paper gives more insight into effects of backtracking in real-time heuristic search. Thus, we hope this will help designing high-performance real-time heuristic search algorithms that take advantage of backtracking.

\section{Conclusions}

Over the last two decades a number of extensions have been implemented to the original RTA*/LRTA* real-time heuristic search algorithm. One of the most radical extensions is backtracking which has been studied primarily empirically. Consequently, the reported trends were highly sensitive to the testbed insomuch as the effects observed in pathfinding on large maps were inconsistent to the effects observed on smaller maps~\cite{Bulitko:05d}.  In this paper we presented the first entirely theoretical analysis of backtracking. In an attempt to make the results as general as possible, we imposed a minimum set of restrictions on the search algorithm. Yet, a tight non-trivial bound was derived on solution cost (exponential in the parameter controlling the amount of backtracking). We then imposed additional restrictions on the search algorithm and showed that they lead to linear solution costs.

\section*{Acknowledgements}
We appreciate feedback from Jonathan Schaeffer, Joe Culberson, Sven Koenig, Rich Korf, Blai Bonet and Hector Geffner. We are grateful for financial support from  the National Science and Engineering Research Council, the University of Alberta, the Alberta Ingenuity Centre for Machine Learning, and the Informatics Circles of Research Excellence.

\appendix

\clearpage\section{Criterion of $\theta$-admissibility}\label{sec:max-min-criterion}

\begin{figure}[ht]
\begin{center}
\includegraphics[width=15cm]{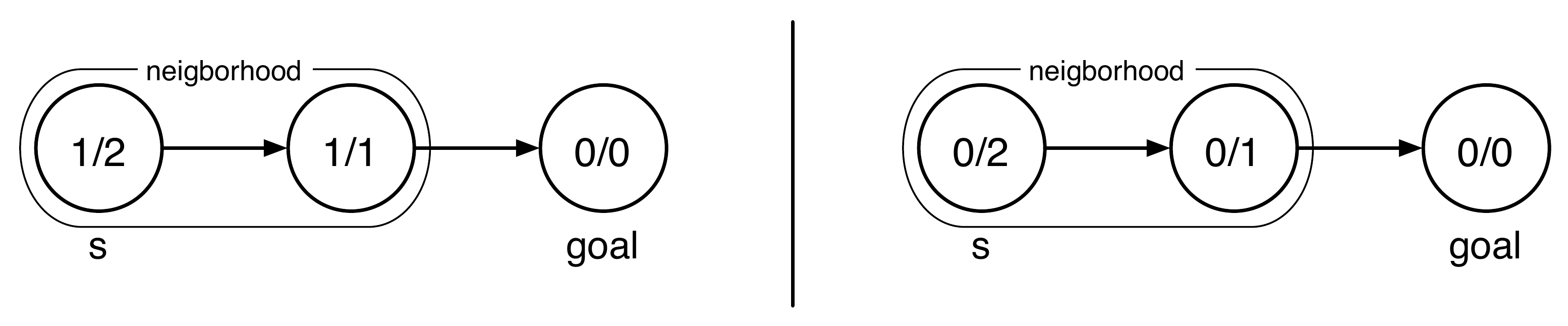}
\vspace{-0.2cm} \caption{{\bf Left:} a three-state  search problem with the current state $s$ and the search neighborhood including the middle state. The heuristic values are shown in the states as $h_t / h^*$. Each edge has the cost of $1$. Increasing the heuristic in state $s$ above the value of 2 dictated by the ``max of min" rule will break $1$-admissibility of the heuristic. {\bf Right:} The same three-state search problem where the value of state $s$ can be increased to $2$, thus exceeding the $1$ dictated by the ``max of min" rule. Yet the $1$-admissibility will be preserved.}\label{fig:c4c4star-naive}
\end{center}\vspace{-0.7cm}
\end{figure}

It is fairly easy to come up with an example  of a policy that does not satisfy the ``max of min" condition and, consequently, does not preserve the $\theta$-admissibility of the heuristic. Remarkably, heuristic updates more aggressive than those dictated by the ``max of min" rule that do preserve $\theta$-admissibility are also possible. An example of each phenomenon is found in Figure~\ref{fig:c4c4star-naive}.

It turns out that the ``max of min"  rule is a sufficient but not necessary condition for preserving $\theta$-admissibility. However, the ``max of min" rule can be strengthened into a necessary condition as follows. Specifically, with a slight modification to the ``max of min" condition based on Definition~\ref{def:raisedHeuristic}, the criterion in Theorem~\ref{th:c4c4star} is proved.

\bdf\label{def:raisedHeuristic} For any  $\Gamma \subset S$, let us define the function $h^\Gamma_t$ as: \bea h^\Gamma_t(s) = \max_{s' \in \Gamma} \left \{ h_t(s) , h_t(s') - \theta \dist(s,s') \right \} \eea where $h_t(s)$ is the heuristic value of state $s$ at time $t$.
The idea underlying $h^\Gamma_t(s)$ is  that if the heuristic $h_t(s)$ is $\theta$-admissible in all states at time $t$, then for any state $s$ it can actually be increased at least to the value of its arbitrary neighbor $s'$ minus the shortest distance from $s$ to $s'$. \edf

\bdf\label{def:c4star} We say that a  search obeys a strengthened ``max of min" condition if at any time $t$: \bea \forall s \in \Gamma^*(\sigma_t) \left [ h_{t+1}(s) \le \max_{J \in \frak D^*(s,\Gamma^*(\sigma_t))} \left (\min_{s' \in J} \left [ \theta \dist(s,s')
+ h_t^{\Gamma^*(\sigma_t)}(s') \right ] \right ) \right ],\label{eq:c4star} \eea where $\Gamma^*(\sigma_t)$ is the union of all neighborhoods considered by the policy up to the time $t$. Formally, $\Gamma^*(\sigma_t) = \cup_{t' \le t} \Gamma(\sigma_t)$. Additionally, $\frak D^*(s,\Gamma) = \frak D(s, \Gamma) \cup
\{\{s\}\}$. The ``raised" heuristic $h_t^{\Gamma^*(\sigma_t)}(s')$ comes from Definition~\ref{def:raisedHeuristic}. \edf

\bt\label{th:c4c4star} {\bf (maximum  heuristic increase criterion).} Suppose we start with $\theta$-admissible heuristic $h$ such that $h(s)=0\iff s\in S_g$. Then as long as $\Gamma^*(\sigma_t)\cap S_g=\O$, the condition of $\theta$-admissibility (Equation~(\ref{eq:c4})) and the strengthened ``max of mins" condition are interchangeable in the definition of $\theta$-admissible policy $\pi$.
\et

\bibliography{arxiv-fta}
\bibliographystyle{theapa}

\end{document}